\documentclass[reprint,aps,prx,superscriptaddress,amsmath,amssymb]{revtex4-2}
\usepackage{graphicx}
\usepackage{newtxtext,newtxmath}

\newcommand{\derivwrt}[2]{\frac{\mathrm{d}#1}{\mathrm{d}#2}}

\begin{document}

\title{Universal Scaling Laws of Absorbing Phase Transitions\\ in Artificial Deep Neural Networks}

\author{Keiichi Tamai}
\email[]{tamai@phys.s.u-tokyo.ac.jp}
\affiliation{Institute for Physics of Intelligence, The University of Tokyo, 7-3-1 Hongo, Bunkyo-ku, Tokyo 113-0033, Japan}
\author{Tsuyoshi Okubo}
\affiliation{Institute for Physics of Intelligence, The University of Tokyo, 7-3-1 Hongo, Bunkyo-ku, Tokyo 113-0033, Japan}
\author{Truong Vinh Truong Duy}
\affiliation{AI Laboratory, Aisin Corporation, 1-18-13 Sotokanda, Chiyoda-ku, Tokyo 101-0021, Japan}
\author{Naotake Natori}
\affiliation{AI Laboratory, Aisin Corporation, 1-18-13 Sotokanda, Chiyoda-ku, Tokyo 101-0021, Japan}
\author{Synge Todo}
\affiliation{Department of Physics, The University of Tokyo, 7-3-1 Hongo, Bunkyo-ku, Tokyo 113-0033, Japan}
\affiliation{Institute for Physics of Intelligence, The University of Tokyo, 7-3-1 Hongo, Bunkyo-ku, Tokyo 113-0033, Japan}
\affiliation{Institute for Solid State Physics, The University of Tokyo, 5-1-5 Kashiwanoha, Kashiwa, Chiba 277-8581, Japan}

\date{\today}

\begin{abstract}
	We demonstrate that conventional artificial deep neural networks operating near the phase boundary of the signal propagation dynamics---also known as the edge of chaos---exhibit universal scaling laws of absorbing phase transitions in non-equilibrium statistical mechanics.
	We exploit the fully deterministic nature of the propagation dynamics to elucidate
	an analogy between a signal collapse in the neural networks and an absorbing state
	(a state that the system can enter but cannot escape from).
	Our numerical results indicate that the multilayer perceptrons and the convolutional neural networks belong to the mean-field and the directed percolation universality classes, respectively.
	Also, the finite-size scaling is successfully applied, suggesting a potential connection to the depth-width trade-off in deep learning.
	Furthermore, our analysis of the training dynamics under the gradient descent reveals that hyperparameter tuning to the phase boundary is necessary but insufficient for achieving optimal generalization in deep networks.
	Remarkably, nonuniversal metric factors associated
	with the scaling laws
	are shown to play a significant role in concretizing the above observations.
	These findings highlight the usefulness of the notion of criticality for analyzing the behavior of artificial deep neural networks and offer new insights toward a unified understanding of the essential relationship between criticality and intelligence.
\end{abstract}

\maketitle

\section{Introduction}\label{sect:intro}
Critical phenomena at second-order phase transitions have long been hypothesized
to be the key to the extraordinary computational power of living systems \cite{Turing1950,Langton1990,Munoz2018}. 
The idea behind the hypothesis is that information cannot propagate through ordered states of matter, 
and it rapidly decays to random noise in the disordered states due to the overly enhanced capability of the medium
to convey disturbance \cite{Langton1990}. Notably, the notion of \textit{phases of matter} lies
at the core of the hypothesis; this viewpoint focuses on collective aspects of the systems,
complementary to more traditional reductionist ones \cite{Bickle2006,Kandel2021}.
Despite the experimental and theoretical challenges stemming from
the many-body nature of the problem, 
pursuing the \textit{computation at the criticality} hypothesis has proven to be 
a fruitful research direction \cite{Munoz2018,Beggs2022}.
In particular, brain dynamics has been intensively discussed \cite{Girardi-Schappo2021,Gros2021}
in the context of absorbing phase transitions \cite{Hinrichsen2000,Henkel2008} 
due to the theoretical relation \cite{Dickman1998}
to self-organized criticality \cite{Bak1987,Hesse2014},
not to mention the straightforward correspondence between death and an absorbing state.

\begin{figure*}[tbp]
	\centering
	\includegraphics[width=\textwidth]{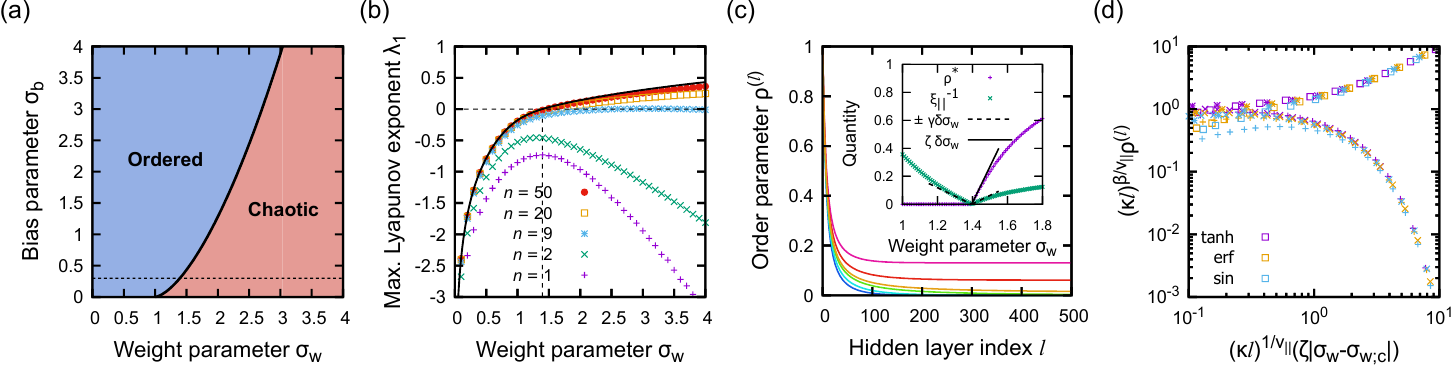}
	\caption{Order-to-chaos transition in the multilayer perceptions
		(\ref{eq:FCrecursive})
		and the associated universal scaling laws.
		(a)~The phase diagram of the signal propagation for $\tanh$ activation.
		The solid curve indicates the phase boundary.
		The dashed line indicates $\sigma_b=0.3$, 
		where the maximum Lyapunov exponent $\lambda_1$ (see Eq.~(\ref{eq:pseudoMLEdef})) and
		the order parameter $\rho^{(l)}$ (see Eq.~(\ref{eq:rho-c-correspondence})) are 
		studied in (b) and (c), respectively.
		(b)~$\lambda_1$ as a function of 
		the weight parameter $\sigma_w$
		for various widths $n$,
		estimated from the convergence of the finite counterpart
		of Eq.~(\ref{eq:pseudoMLEdef}) at $l=10^5$.
		The vertical line indicates the position of the critical point ($\sigma_{w;c}\sim 1.39558$).
		The black curve is the result expected in the limit of $n\rightarrow\infty$,
		namely $\lambda_{\mathcal{C}}/2$ (see Eq.~(\ref{eq:CmapLyapunov})).
		(c)~The main panel shows $\rho^{(l)}$ as a function of $l$ for various $\sigma_w$
		(from 1.35 (blue; the ordered phase) to 1.45 (magenta; the chaotic phase)),
		calculated from the mean-field theory (\ref{eq:FCMF-q}) and (\ref{eq:FCMF-c}). 
		The inset shows the stationary value $\rho^*$
		of the order parameter and the reciprocal correlation depth $\xi_{\|}^{-1}$ 
		(see Eqs.\ (\ref{eq:FCcorrdepthdef}) and (\ref{eq:infFCcorrdepth}))
		as a function of $\sigma_w$.
		The dashed line and the solid line are guides-to-eye for linear onset with
		a slope of $\gamma_{\leftrightarrow}$ and
		$\zeta_{\leftrightarrow}:=\gamma_{\leftrightarrow}/\kappa^{1/\nu_{\|}}$, respectively
		($\delta\sigma_w:=\sigma_{w}-\sigma_{w;c}$).
		(d)~The order parameter $\rho^{(l)}$ as a function of $l$ for various activation functions
		is rescaled according to the universal scaling ansatz (\ref{eq:abs-ansatz}).
		Different symbols with the same color correspond to
		different values of $\sigma_w$:
		$+$, $\times$, $*$, $\square$ in ascending order ($\sigma_b$ is fixed to be 0.3).
		In (c) and (d), the nonuniversal metric factors $\gamma_{\leftrightarrow}$ and $\kappa$ are
		calculated from
		Eqs.\ (\ref{eq:gamma-formula}) and (\ref{eq:kappa-formula}), respectively, for each case.
	}
	\label{fig:DNNAPTanalogy}
\end{figure*}

The rapid progress in applying deep learning techniques \cite{Silver2016,Stahlberg2020,Rombach2022}
motivates us to ask whether artificial deep neural networks
also utilize criticality for their performance.
Recent theoretical studies on infinitely wide networks suggest this is the case.
Under a specific setup (see also Sect.~\ref{sect:setups}), the signal propagation dynamics
in untrained deep neural networks can be classified into two phases: 
the ordered phase and the chaotic phase, depending on the hyperparameters 
used for initialization \cite{Poole2016,Schoenholz2017};
see also Fig.~\ref{fig:DNNAPTanalogy}(a).
The network with sufficiently many hidden layers
returns almost the same outputs for any inputs in the ordered phase,
whereas decorrelated outputs in the chaotic phase.
In either case, the network cannot remember the degree of similarity between different inputs,
which limits the performance as a learning agent.
As such, the network at the ordered phase suffers from vanishing gradient and untrainability,
while at the chaotic phase from exploding gradient and ungeneralizability \cite{Schoenholz2017,Xiao2020}.
Consequently, the phase boundary, often called \textit{the edge of chaos},
attracts considerable interest in deep learning research,
even though some studies indicate that initialization at the edge
alone does not necessarily lead to good generalization \cite{Peluchetti2020,Hayou2022}.
Remarkably, the characteristic depth of the network dynamics is
suggested to diverge at the edge of chaos \cite{Schoenholz2017},
highly reminiscent of critical phenomena at second-order phase transitions.
Subsequent works extend similar results 
to various activation functions \cite{Hayou2019,Roberts2022} 
and network architectures \cite{Xiao2018,Yang2017}.

From a statistical mechanics viewpoint, universal scaling laws,
if any, are relevant for characterizing the critical phenomena
at the edge of chaos in deep neural networks. 
Even though the mean-field theory in a suitable limit \cite{Poole2016,Xiao2018}
and its perturbative expansion \cite{Grosvenor2022,Yaida2020} 
are available in simple cases, 
the scaling laws may provide complementary, flexible, 
and powerful insight into the network dynamics:
thanks to the universality of the critical phenomena, 
intuitive phenomenological considerations 
may result in at least partially quantitative predictions.
Also, theoretical tools such as finite-size \cite{Ferdinand1969} 
or short-time \cite{Janssen1989} scaling 
shed further light on the dynamics beyond the limiting cases.
Besides the benefits for our understanding of deep neural networks,
embedding them better in statistical mechanics
may provide clues for studying how living systems perform intellectual tasks.

Below, we demonstrate the connection between deep neural network dynamics 
at the edge of chaos and absorbing phase transitions \cite{Hinrichsen2000,Henkel2008}.
After some preliminaries (Sect.~\ref{sect:setups}), 
we establish a correspondence between the ordered state of deep neural networks
and an absorbing state by studying the linear stability
of the former (Sect.~\ref{sect:linear-stability}).
Next, in the case of the multilayer perceptrons, we thoroughly investigate the scaling properties
of the phase transition
in the thermodynamic limit (Sect.~\ref{subsec:MFreview} and \ref{subsec:infwide-scaling}). 
Remarkably, the scaling properties of the neural tangent kernel (NTK) \cite{Jacot2018}
provide a novel insight into
the \textit{curse of depth} reported in the literature \cite{Hayou2022}. 
The investigation is then extended to
finite width and different architectures (Sect.~\ref{subsec:scaling-finite-and-arch}),
although we content ourselves only with front propagation dynamics in these cases. 
In particular, we provide numerical evidence of the directed percolation universality
in convolutional neural networks. We conclude the paper with a brief discussion
on possible directions for future work (Sect.~\ref{sect:discussion}).

\newcommand{\qfixedpt}{q^*}
\newcommand{\cfixedpt}{c^*}
\newcommand{\stdnormpdf}{\frac{1}{\sqrt{2\pi}}e^{-\frac{z^2}{2}}}
\newcommand{\zstdnormpdf}{\frac{z}{\sqrt{2\pi}}e^{-\frac{z^2}{2}}}
\section{Preliminaries}\label{sect:setups}
To illustrate our view with a simple setup,
we exclusively consider the multilayer perceptrons
with the NTK parameterization \cite{Jacot2018} until Sect.~\ref{subsec:infwide-scaling}.
Formally, the recurrence relation for the preactivation $\boldsymbol{z}^{(l)}$
at the $l$-th hidden layer ($l=1,2,\cdots,L$, where $L$ is the depth of the network) 
and the output $y$, assumed to be of a single element for simplicity, 
are written as follows:
\begin{align}
	\label{eq:FCrecursive}
	z_i^{(l+1)}&=
	\begin{cases}
		\displaystyle \frac{\sigma_w}{\sqrt{n_{\mathrm{in}}}}\sum_{j}W_{ij}^{(l+1)}x_j+\sigma_bb_i^{(l+1)} &  l=0;\\
		\displaystyle \frac{\sigma_w}{\sqrt{n}}\sum_{j}W_{ij}^{(l+1)}h(z^{(l)}_j)+\sigma_bb_i^{(l+1)} &  1\le l < L,
	\end{cases}
	\\
	y&=\frac{\sigma_w}{\sqrt{n}}\sum_{j} W^{(L+1)}_{1j}h(z_j^{(L)})+\sigma_b b^{(L+1)}.
\end{align}
Here, $\boldsymbol{x}$ is the input with $n_{\mathrm{in}}$ elements, 
$W^{(l)}$ and $\boldsymbol{b}^{(l)}$ the weight matrix 
and the bias vector at the $l$-th hidden layer, respectively,
$\sigma_w,\sigma_b$ the associated hyperparameters and $n$ the width of the hidden layers;
the meanings of the symbols associated with the neural networks
are also summarized in Table \ref{tab:NNglossary} for convenience.
Every element of the weight and the bias is initialized according to
the standard normal distribution $\mathcal{N}(0,1)$.
This parameterization differs slightly from
the one commonly used in practice \cite{SohlDickstein2020},
but the difference is not essential for our study \cite{Lee2019}.
The activation function $h$ is assumed to be 
in the $K^*=0$ universality class
in the sense of Roberts \textit{et al.}\ \cite{Roberts2022},
a few examples being $\tanh$, $\mathrm{erf}$, and $\sin$, but not $\mathrm{ReLU}$
(note, however, that similar observations can be made
also for ReLU-like activation functions;
see Appendix~\ref{sect:scale-invariant-activation}).

Since the forward dynamics (\ref{eq:FCrecursive}) is fully deterministic after initialization,
a pair of signals may end up with a collapse:
once the two signals $\boldsymbol{z}_1^{(l)}, \boldsymbol{z}_2^{(l)}$
become identical at one hidden layer,
they never deviate from each other again at deeper hidden layers.
Interpreting the network depth as a temporal degree of freedom,
the ordered state $\boldsymbol{z}_1^{(l)}=\boldsymbol{z}_2^{(l)}$ can be
regarded as an absorbing state of the dynamics.
Even though the exact order rarely occurs in practice 
because it requires an accidentally degenerate weight matrix,
the difference between two signals can decay exponentially in some cases,
especially when the network is narrow.
Meanwhile, with sufficient width and a suitable choice of the hyperparameters $(\sigma_w,\sigma_b)$, 
one can observe the opposite: magnification of the difference, even if initially tiny.
In this case, the difference no longer converges to a unique $l\rightarrow\infty$ limit
but instead fluctuates randomly. Thus, the networks seem to exhibit
a phase transition between a unique absorbing (ordered) phase
and a fluctuating active (chaotic) phase.
As we will see shortly,
the transition can be further formalized by studying
the linear stability of the exactly ordered state.

\begin{table}
	\caption{Glossary of the symbols associated with the neural networks. If accompanied by the superscript ``$(l)$,''
		 the symbol indicates the corresponding quantity at the $l$-th hidden layer.}
	\label{tab:NNglossary}
	\begin{tabular}{c|c}
		\hline
		Symbol & Meaning\\
		\hline
		$\boldsymbol{x}$ & Input vectors.\\
		$W$ & Weight matrices.\\
		$\boldsymbol{b}$ & Bias vectors.\\
		$\sigma_w,\sigma_b$ & Hyperparameters associated with the network,\\
		& corresponding to standard deviation of the weight/bias.\\
		$n_{\text{in}}$ & The number of the elements of the input.\\
		$n$ & Width of the network.\\
		$L$ & Depth of the network.\\
		$\boldsymbol{z}$ & Preactivation: the signal vector just before\\ 
		 & applying the activation function.\\
		$h$ & Activation function.\\
		$q$ & Preactivation variance in the mean-field theory.\\
		$C$ & Preactivation covariance in the mean-field theory.\\
		\hline
	\end{tabular}
\end{table}

\section{Absorbing property of\\the ordered state}\label{sect:linear-stability}
To address the issue of the linear stability of the ordered state, we study
(with a slight abuse of language) the maximum Lyapunov exponent
for the front propagation dynamics (\ref{eq:FCrecursive}):
\begin{equation}
	\label{eq:pseudoMLEdef}
	\lambda_1:=\lim_{l\rightarrow\infty}\frac{1}{l}\log\frac{\|J^{(l+1)}(\boldsymbol{z}^{(l)})\cdots J^{(2)}(\boldsymbol{z}^{(1)})\boldsymbol{u}_0\|_2}{\|\boldsymbol{u}_0\|_2}, 
\end{equation}
where $\boldsymbol{u}_0\in\mathbb{R}^{n}$ is an arbitrary nonzero vector and $J^{(l)}$ is the layer-wise input-output Jacobian
\begin{equation}
	J^{(l)}(\boldsymbol{z})=\frac{\sigma_w}{\sqrt{n}}\left(
	\begin{array}{ccc}
		J_{11}^{(l)}(\boldsymbol{z}) & \cdots & J_{1n}^{(l)}(\boldsymbol{z})\\
		\vdots & \ddots & \vdots\\
		J_{n1}^{(l)}(\boldsymbol{z}) & \cdots & J_{nn}^{(l)}(\boldsymbol{z})\\
	\end{array}
	\right)
\end{equation}
with \footnote{The standard Lagrange's (prime) notation
	for differentiation is used throughout the paper.}
\begin{equation}
	J_{ij}^{(l)}(\boldsymbol{z}):=W_{ij}^{(l)}h^{\prime}(z_j).
\end{equation}
By doing so, we can directly see how the notion of
the order-to-chaos transition emerges
as a many-body effect in the neural networks.
For instance, $\lambda_1$ as a function of $\sigma_w$
for $\tanh$ activation function is shown in Fig.~\ref{fig:DNNAPTanalogy}(b).
In this case, $\lambda_1$ is negative in the entire domain for small width $n$,
which suggests that the ordered state is always stable against infinitesimal discrepancy.
However, $\lambda_1$ increases as $n$ becomes larger, 
and eventually, $\lambda_1$ changes its sign at some $\sigma_w$ for large $n$,
indicating loss of linear stability.
Naturally, the position of the onset of the linear instability is very close to
that of the critical point predicted from the mean-field theory \cite{Poole2016}
when $n$ is large and is expected to coincide with the limit of $n\rightarrow\infty$. 
One can also empirically see
that the loss of the stability of the ordered state 
at some $\sigma_w$ for sufficiently large $n$ is a robust feature
of the neural networks under the current consideration,
although how $\lambda_1$ as a function of $\sigma_w$ behaves for small $n$
is somewhat more sensitive to the choice of $h$.

Thus, the maximum Lyapunov exponent $\lambda_1$ successfully captures the well-defined transition
from the ordered phase to the chaotic phase, even for finite networks. 
In the ordered phase, once a pair of preactivations $(\boldsymbol{z}_1,\boldsymbol{z}_2)$ reach
reasonably close to the ordered state, they are hard to escape from it. Meanwhile, in the chaotic phase,
a pair of preactivations are allowed to get away from the vicinity of the ordered state,
although the ordered state itself is still absorbing.
This scenario, a transition from a non-fluctuating absorbing phase to a fluctuating active phase,
is highly reminiscent of an absorbing phase transition in statistical mechanics.

\section{Universal scaling around\\the order-to-chaos transition}
Having seen that the order-to-chaos transition is at least conceptually analogous to absorbing phase transitions,
the next step is to seek a deeper connection between these two by further quantitative characterization.

\subsection{Mean-field theory of signal propagation}\label{subsec:MFreview}
The phase transition between the ordered and the chaotic phases
can be quantitatively studied in the limit of wide networks.
In this limit, the neural network becomes equivalent to 
a Gaussian process \cite{Neal1996,Williams1996,Lee2018,Matthews2018},
whose diagonal $q^{(l)}$ and non-diagonal $C^{(l)}$ elements of the covariance matrix
for each hidden layer can be recursively described by the mean-field theory \cite{Poole2016}:
\begin{align}
	\label{eq:FCMF-q}
	q_i^{(l+1)}&=\sigma_w^2\int\mathcal{D}z\,h^2(\sqrt{q_i^{(l)}}z)+\sigma_b^2; \\
	\label{eq:FCMF-c}
	C^{(l+1)}&=\sigma_w^2\int\mathcal{D}z_1\int\mathcal{D}z_2\,h(u_1^{(l)})h(u_2^{(l)})+\sigma_b^2; \\
	\label{eq:auxUdef}
	\begin{split}
		\displaystyle u_1^{(l)}&:=\sqrt{q_1^{(l)}}z_1;\\
		\displaystyle u_2^{(l)}&:=\sqrt{q_2^{(l)}}\left(c^{(l)}z_1+\sqrt{1-(c^{(l)})^2}z_2\right)
	\end{split}
\end{align}
with the initial conditions
\begin{equation}
	\label{eq:FCMF-initconds}
	q_i^{(1)}=\sigma_w^2\frac{\|\boldsymbol{x}_i\|_2^2}{n_{\mathrm{in}}}+\sigma_b^2,\quad C^{(1)}=\sigma_w^2\frac{\boldsymbol{x}_1\cdot\boldsymbol{x}_2}{n_{\mathrm{in}}}+\sigma_b^2,
\end{equation}
where $i=1,2$,
\begin{equation}
	\label{eq:MFPearsoncorr}
	c^{(l)}:=\frac{C^{(l)}}{\sqrt{q_{1}^{(l)}q_2^{(l)}}}
\end{equation}
is the Pearson correlation coefficient, and
\begin{equation}
	\int\mathcal{D}z:=\int_{-\infty}^{\infty}\mathrm{d}z\frac{1}{\sqrt{2\pi}}e^{-\frac{z^2}{2}}.
\end{equation}

Let us recall some basic results of this theory.
One can see that $q^{(l)}$ rapidly converges to a fixed point $q^*:=\lim_{l\rightarrow\infty} q^{(l)}$
as the depth $l$ tends to infinity \cite{Poole2016,Schoenholz2017}, generally without a sign of a phase transition 
(unless $\sigma_b=0$, where $q^*$ vanishes at the ordered phase). 
By substituting $q_1^{(l)}=q_2^{(l)}=q^*$ to Eqs.~(\ref{eq:FCMF-c}), (\ref{eq:auxUdef}) and
observing Eq.~(\ref{eq:MFPearsoncorr}), we obtain 
an approximate closed-form description for $c^{(l)}$
also known as the iterative $\mathcal{C}$-map, valid for large $l$:
\begin{equation}
	\label{eq:iterCmap}
	c^{(l+1)}=\frac{1}{q^*}\left[\sigma_w^2\int\mathcal{D}z_1\int\mathcal{D}z_2\,h(u_1^{*(l)})h(u_2^{*(l)})+\sigma_b^2\right],
\end{equation}
where $u_1^{*(l)}:=\sqrt{q^{*}}z_1,u_2^{*(l)}:=\sqrt{q^{*}}(c^{(l)}z_1+\sqrt{1-(c^{(l)})^2}z_2)$.
Linear stability of the trivial fixed point $c^{(l)}=c^{(l+1)}=1$ of Eq.~(\ref{eq:iterCmap}) determines the phase
for a given pair of hyperparameters $(\sigma_w,\sigma_b)$, as depicted in Fig.~1(a):
stable at the ordered phase, whereas unstable at the chaotic phase.
It can be shown that, for a fixed $\sigma_b$,
the discrepancy from the fixed point $c^*$ asymptotically
decays exponentially with a suitable correlation depth $\xi_{\|}$ \cite{Schoenholz2017}:
\begin{equation}
	\label{eq:FCcorrdepthdef}
	\lim_{l\rightarrow\infty}\frac{\log |c^{(l)}-\cfixedpt{}|}{l}=-\frac{1}{\xi_{\|}}
\end{equation}
with $\xi_{\|}$ given by the following:
\begin{equation}
	\label{eq:infFCcorrdepth}
	e^{-\frac{1}{\xi_{\|}}} = 
	\begin{cases}
		\displaystyle \sigma_w^2\int\mathcal{D}z\,h^{\prime2}(\sqrt{\qfixedpt{}}z) & \sigma_w<\sigma_{w;c};\\
		\displaystyle \sigma_w^2\int\mathcal{D}z_1\int\mathcal{D}z_2\,h^{\prime}(u_1^*)h^{\prime}(u_2^*) & \sigma_w>\sigma_{w;c},
	\end{cases}
\end{equation}
where $\sigma_{w;c}$ is the critical point for the specified $\sigma_b$, satisfying
\begin{equation}
	\sigma_{w;c}^2\int\mathcal{D}z\,h^{\prime 2}(\sqrt{q^*}z) = 1.
\end{equation}
Note also that half of the maximum Lyapunov exponent
for the trivial fixed point of the iterative $\mathcal{C}$-map
\begin{equation}
	\label{eq:CmapLyapunov}
	\lambda_{\mathcal{C}}:=\log\left(\sigma_w^2\int\mathcal{D}z\,h^{\prime2}(\sqrt{\qfixedpt{}}z)\right)
\end{equation}
equals to the maximum Lyapunov exponent $\lambda_1$
of the ordered state (see Eq.~(\ref{eq:pseudoMLEdef})) in the limit of infinite width $n$.
Encouragingly, the behavior of $\lambda_1$ for $n=50$
as a function of $\sigma_w$ is
already close to the $n\rightarrow\infty$ limit (Fig.~\ref{fig:DNNAPTanalogy}(b));
this suggests that the infinitely wide neural network may serve
as a good starting point for understanding
the behavior of the neural networks of practical width.

\subsection{Scaling results for the infinitely wide networks}\label{subsec:infwide-scaling}
One of the most common strategies for studying systems with
absorbing phase transition is to examine universal scaling laws \cite{Henkel2008,Hinrichsen2000}.
For instance, the time evolution of the order parameter $\rho(t)$ in the thermodynamic limit
admits the following scaling ansatz: 
\begin{equation}
	\label{eq:abs-ansatz}
	\rho(t;\tau) \sim (\kappa t)^{-\beta/\nu_{\|}} f((\kappa t)^{1/\nu_{\|}} \zeta\tau),
\end{equation}
where $\tau$ denotes the discrepancy from the critical point, 
$\beta,\nu_{\|}$ are the critical exponents associated
with the onset of order parameter $\rho$
and the correlation time $\xi_{\|}$ of the steady state, respectively, that is,
\begin{equation}
	\label{eq:steady-state-scaling}
	\rho(t\rightarrow\infty)\sim (\zeta\tau)^{\beta},\quad \xi_{\|}\sim |\gamma\tau|^{-\nu_{\|}} \quad \mathrm{as}\quad \tau\rightarrow 0;
\end{equation}
\begin{equation}
	\label{eq:gamma-zetarelation}
	\gamma := \zeta\kappa^{1/\nu_{\|}}.
\end{equation}
Remarkably, the critical exponents and the scaling function $f$
are the same for all systems in a given universality class, while the specific details are summarized
in the nonuniversal metric factors $\kappa,\zeta,\gamma$ \cite{Privman1984},
two of which are independent. 

Let us investigate the universal scaling laws in the signal propagation dynamics.
In the present context, we define the order parameter $\rho$ to be
the Pearson correlation coefficient between preactivations for different inputs, which is then
subtracted from unity so that $\rho$ vanishes in the ordered phase.
In particular, when the network is infinitely wide,
the order parameter $\rho^{(l)}$ at each hidden layer is
directly related to $c^{(l)}$ (see Eq.~(\ref{eq:MFPearsoncorr})) in the mean-field theory:
\begin{equation}
	\label{eq:rho-c-correspondence}
	\rho^{(l)}:=1-c^{(l)}.
\end{equation}

We can show that the multilayer perceptrons
exhibit the universal scaling laws identical to those of
the mean-field theory for absorbing phase transitions \cite{Henkel2008,Hinrichsen2000}.
Specifically, $\rho^{(l)}$ and the correlation depth $\xi_{\|}$ (see Eq.~(\ref{eq:FCcorrdepthdef})) 
exhibits the power-law scaling (\ref{eq:steady-state-scaling}) with
\begin{equation}
	\label{eq:FCcrit-exponents}
	\beta=1,\quad \nu_{\|}=1.
\end{equation}
This can be shown by considering an infinitesimally small deviation
from the critical point, and expand the mean-field theory (\ref{eq:FCMF-q}), (\ref{eq:FCMF-c})
to track the change of the position of the fixed point
and of the correlation depth (\ref{eq:infFCcorrdepth}) up to
the lowest relevant order of the deviation;
see Appendix \ref{sect:FCexponentderivation} for details.
An interesting corollary is that the nonuniversal metric factors $\kappa,\gamma$
in the sense of Eq.~(\ref{eq:abs-ansatz}) can be evaluated theoretically
in the present case. For instance, if we choose to fix
the bias parameter $\sigma_b$ and vary the weight parameter $\sigma_w$
(the discrepancy $\tau$ from the critical point is defined to be $\sigma_w-\sigma_{w;c}$),
we find
\begin{equation}
	\label{eq:gamma-formula}
	\gamma_{\leftrightarrow} = \frac{2}{\sigma_{w;c}}\left(1-\frac{\displaystyle (q_c^*-\sigma_b^2)\int\mathcal{D}z\,zh^{\prime}(\sqrt{q_c^*}z)h^{\prime\prime}(\sqrt{q_c^*}z)}{\displaystyle \sqrt{q_c^*}\int\mathcal{D}z\,h(\sqrt{q^*_c}z)h^{\prime\prime}(\sqrt{q^*_c}z)}\right),
\end{equation}
\begin{equation}
	\label{eq:kappa-formula}
	\kappa = \frac{\displaystyle q^*_c\int\mathcal{D}z\,h^{\prime\prime 2}(\sqrt{q^*_c}z)}{\displaystyle 2\int\mathcal{D}z\,h^{\prime 2}(\sqrt{q^*_c}z)},
\end{equation}
where $q^*_c$ is the fixed point of Eq.~(\ref{eq:FCMF-q})
at the critical point
and the arrow symbol $\leftrightarrow$ indicates the direction
in which we cross the boundary in the phase diagram (Fig.~\ref{fig:DNNAPTanalogy}(a)).
We empirically validate the results
in Fig.~\ref{fig:DNNAPTanalogy}(c) and Fig.~\ref{fig:DNNAPTanalogy}(d).
In particular, we see that the order parameter dynamics $\rho^{(l)}$ for 
various activation functions collapse into 
a single universal curve, as predicted 
by the scaling ansatz~(\ref{eq:abs-ansatz}),
except when $l$ is small just as expected.
These observations further support the view that
the network exhibits an absorbing phase transition
into the ordered state.
Similar results
(albeit with different metric factor $\gamma_{\updownarrow}$) can be obtained
if $\sigma_w$ is fixed and $\sigma_b$ is varied,
provided that $\sigma_w > (h^{\prime}(0))^{-1}$:
\begin{equation}
	\label{eq:metricgamma-fixedw}
	\gamma_{\updownarrow} = \frac{\displaystyle 2\sigma_{b;c}\int\mathcal{D}z\,zh^{\prime}(\sqrt{q_c^*}z)h^{\prime\prime}(\sqrt{q_c^*}z)}{\displaystyle \sqrt{q_c^*}\int\mathcal{D}z\,h(\sqrt{q_c^*}z)h^{\prime\prime}(\sqrt{q_c^*}z)},
\end{equation}
where $\sigma_{b;c}$ is the critical point for the specified $\sigma_w$.

\begin{figure}[tbp]
	\centering
	\includegraphics[width=\columnwidth]{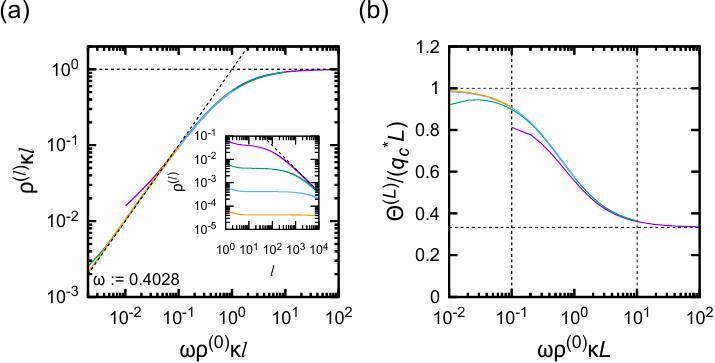}
	\caption{Critical dynamic scaling of the order parameter $\rho^{(l)}$ and its consequence on
		neural tangent kernel (NTK). (a)~The order parameter $\rho^{(l)}$ at a critical point 
		(specifically $(\sigma_w,\sigma_b)\sim(1.23367,0.3)$ with erf activation;
		$\kappa\sim 0.252674$)
		for various cosine distances $\rho^{(0)}$
		(ranging from $10^{-4}$ (orange) to $10^{-1}$ (purple); see Eq.~(\ref{eq:cosdist-def})),
		calculated from the mean-field theory (\ref{eq:FCMF-q}), (\ref{eq:FCMF-c}) and
		rescaled according to the scaling ansatz (\ref{eq:initial-slip-ansatz}).
		The inputs $\boldsymbol{x}_1,\boldsymbol{x}_2$ with $n_{\text{in}}=10$ elements
		were normalized so that $\|\boldsymbol{x}_1\|_2=\|\boldsymbol{x}_2\|_2=1$.
		The dashed lines are guides-to-eye for 
		the asymptotic behavior (\ref{eq:slipg-asymptotes})
		of the scaling function $g$. The raw $\rho^{(l)}$ is
		shown in the inset with a guide to eye for $\rho^{(l)}=(\kappa l)^{-1}$.
		(b)~The NTK $\Theta^{(L)}(\boldsymbol{x}_1,\boldsymbol{x}_2)$ (see Eq.~(\ref{eq:FC-NTK}))
		as a function of the network depth $L$,
		rescaled according to the scaling ansatz (\ref{eq:NTK-ansatz}).
		The same $(\sigma_w,\sigma_b,\rho^{(0)})$ and activation function
		as (a) were used.
		The two horizontal lines are guides-to-eye for
		the asymptotic behavior (\ref{eq:NTKasymptotes}) as $L\rightarrow\infty$:
		the upper one for $\boldsymbol{x}_1=\boldsymbol{x}_2$ ($\rho^{(0)}=0$),
		the lower one for $\boldsymbol{x}_1\ne\boldsymbol{x}_2$ ($\rho^{(0)}>0$).
	}
	\label{fig:DNNAPTscaling}
\end{figure}

The nonuniversal metric factor $\kappa$ deserves special attention
because it serves as an intrinsic characterizer of a critical point.
That is, $\kappa$ is uniquely determined once a critical point is specified,
in contrast with $\gamma$ and $\zeta$, which also depend on how we approach the critical point.
Formally $\kappa$ is the reciprocal amplitude of the power-law decay at a critical point
\begin{equation}
	\rho^{(l)}\sim (\kappa l)^{-\beta/\nu_{\|}}\quad \mathrm{for}\quad l\gg 1,
\end{equation} 
but the readers might ask for a more intuitive meaning.
The scaling laws for a critical initial slip \cite{Janssen1989}
can be utilized to address this question. 
At a critical point, 
we find that the order parameter $\rho^{(l)}$
for various cosine distances
\begin{equation}
	\label{eq:cosdist-def}
	\rho^{(0)}:=1-\frac{\boldsymbol{x}_1\cdot \boldsymbol{x}_2}{\|\boldsymbol{x}_1\|_2\|\boldsymbol{x}_2\|_2}
\end{equation}
of the normalized inputs $\boldsymbol{x}_1,\boldsymbol{x}_2$ exhibits
the universal scaling law described by the following scaling ansatz
(Fig.~\ref{fig:DNNAPTscaling}(a)):
\begin{equation}
	\label{eq:initial-slip-ansatz}
	\rho^{(l)}\simeq (\kappa l)^{-1}g(\omega\rho^{(0)}\kappa l),
\end{equation}
where $\omega$ is a metric factor associated with an initial condition
depending on a critical point of interest and the inputs
\footnote{In particular, if we normalize the input so that $q^{(1)}$ matches
	the fixed point $q^*$ of the mean-field theory (\ref{eq:FCMF-q}), $\omega$ equals to
	the proportionality constant between $\rho^{(0)}$ and
	the initial condition $\rho^{(1)}$ of the mean-field theory: $\rho^{(1)}=\omega \rho^{(0)}$
	with $\omega = \sigma_w^2q^*/(\sigma_w^2q^*+n_{\text{in}}\sigma_b^2)$.}.
Notably, the scaling function $g$ shows a crossover between two asymptotic behaviors:
\begin{equation}
	\label{eq:slipg-asymptotes}
	g(x)\sim
	\begin{cases}
		x & x\ll 1;\\
		1 & x\gg 1.
	\end{cases}
\end{equation}
Which asymptotic regime the signal propagation dynamics belongs to is
a matter of comparison between $\rho^{(0)}$ and $(\omega\kappa l)^{-1}$.
This suggests a striking resemblance to
the cosine distance scoring~\cite{Dehak2011},
where a simple thresholding on the cosine distance of the feature vectors
yields fast and robust speaker verification.
In the case of the multilayer perceptrons, the threshold for the crossover is
determined implicitly by specifying a critical point ($\kappa,\omega$),
the depth of the network ($l$), and how we design the inputs ($\omega$).
In other words, the metric factor $\kappa$, \textit{combined with the depth},
characterizes the network's sensitivity against input differences.

To gain a deeper insight into the implications of the scaling laws
for the training dynamics of the neural networks,
let us study the neural tangent kernel (NTK) \cite{Jacot2018} 
\begin{equation}
	\label{eq:general-NTK}
	\begin{array}{l}
		\displaystyle \Theta^{(L)}(\boldsymbol{x}_1,\boldsymbol{x}_2;\boldsymbol{\theta}_0) := \sum_j\frac{\partial y}{\partial \theta_j}(\boldsymbol{x}_1;\boldsymbol{\theta}_0) \frac{\partial y}{\partial\theta_j}(\boldsymbol{x}_2;\boldsymbol{\theta}_0)\\
	\end{array}
\end{equation}
of the initialized networks
(here, $y(\boldsymbol{x};\boldsymbol{\theta})$ is 
the output and $\boldsymbol{\theta}_0$ is an initial value of the trainable parameters,
namely $\{W^{(l)},\boldsymbol{b}^{(l)}\}_{l=1}^{L+1}$).
In the limit of infinite width,
the initial value of NTK is deterministic
despite the randomness of $\boldsymbol{\theta}_0$ themselves,
and also NTK stays constant during
the training under the gradient descent using mean squared error
with small learning rate \cite{Lee2019}.
Consequently, the dynamics of the output $y(\boldsymbol{x};\boldsymbol{\theta})$
under such circumstances can be reduced to
a linear ordinary differential equation.
In particular, a collection of the residual errors $\Delta\boldsymbol{y}(\boldsymbol{\theta}):=(\Delta y(\boldsymbol{x}_1;\boldsymbol{\theta}),\cdots,\Delta y(\boldsymbol{x}_N;\boldsymbol{\theta}))^T$
for training inputs
$\{\boldsymbol{x}_1,\cdots,\boldsymbol{x}_N\}$
is governed by
\begin{equation}
	\frac{\mathrm{d}\Delta \boldsymbol{y}(\boldsymbol{\theta}(t))}{\mathrm{d}t} = -\frac{2\eta}{N}\Theta_{\mathrm{train}}^{(L)}\Delta\boldsymbol{y}(\boldsymbol{\theta}(t)),
\end{equation}
where $\eta$ is a learning rate and $\Theta_{\mathrm{train}}^{(L)}$ is
the following matrix (the explicit $\boldsymbol{\theta}_0$-dependence is
dropped due to the deterministic property):
\begin{equation}
	\Theta_{\mathrm{train}}^{(L)}:=\left(
	\begin{array}{ccc}
		\Theta^{(L)}(\boldsymbol{x}_1,\boldsymbol{x}_1) & \cdots & \Theta^{(L)}(\boldsymbol{x}_1,\boldsymbol{x}_N)\\
		\vdots & \ddots & \vdots\\
		\Theta^{(L)}(\boldsymbol{x}_N,\boldsymbol{x}_1) & \cdots & \Theta^{(L)}(\boldsymbol{x}_N,\boldsymbol{x}_N)\\
	\end{array}
	\right).
\end{equation}
NTK also plays a significant role in characterizing generalizability of
the infinitely-wide neural networks trained under the stochastic gradient descent \cite{Cao2019}.
Thus, the initial value of NTK is relevant for understanding the training dynamics
and performance of the neural networks.

The connection between the initialized NTK and the universal scaling laws can be seen by
observing that the closed-form expression \cite{Arora2019} of the NTK
for the present case is described in terms of $u_1^{(l)},u_2^{(l)}$, and $C^{(l)}$
in the mean-field theory:
\begin{equation}
	\label{eq:FC-NTK}
	\begin{array}{l}
		\displaystyle \Theta^{(L)}(\boldsymbol{x}_1,\boldsymbol{x}_2) = \\ \displaystyle \sum_{l=1}^{L+1}C^{(l)}\prod_{l^{\prime}=l}^{L} \left(\sigma_w^2\int\mathcal{D}z_1\int\mathcal{D}z_2\,h^{\prime}(u_1^{(l^{\prime})})h^{\prime}(u_2^{(l^{\prime})})\right).\\
	\end{array}
\end{equation}
The both terms in the RHS of Eq.~(\ref{eq:FC-NTK}) at a critical point
is directly related to the order parameter $\rho^{(l)}$
(to find Eq.~(\ref{eq:derivprod-rho-correspondence}),
perform a similar infinitesimal expansion
to the one demonstrated in Appendix \ref{sect:FCexponentderivation})
\begin{equation}
	\label{eq:C-rho-correspondence}
	C^{(l)} \simeq q_c^{*}(1-\rho^{(l)}),
\end{equation}
\begin{equation}
	\label{eq:derivprod-rho-correspondence}
	\sigma_{w;c}^2 \int\mathcal{D}z_1\int\mathcal{D}z_2\,h^{\prime}(u_1^{(l)})h^{\prime}(u_2^{(l)}) \simeq 1-2\kappa\rho^{(l)}.
\end{equation}
As such, we naturally expect a universal scaling ansatz
akin to Eq.~(\ref{eq:initial-slip-ansatz})
for the NTK at a critical point:
\begin{equation}
	\label{eq:NTK-ansatz}
	\Theta^{(L)}(\boldsymbol{x}_1,\boldsymbol{x}_2) \simeq q_c^{*}L\tilde{g}(\omega\rho^{(0)}\kappa L)
\end{equation}
with a suitable scaling function $\tilde{g}$. Here, the prefactor $q_c^*L$ reflects
the asymptotic proportionality to $L$ established in the literature~\cite{Xiao2020,Hayou2022};
the limits of the scaling function $\tilde{g}$ can also be deduced from the prior works
\begin{equation}
	\label{eq:NTKscalingfunc}
	\tilde{g}(x) \rightarrow
	\begin{cases}
		1 & \text{as}\quad x\rightarrow 0;\\
		1/3 & \text{as}\quad x\rightarrow \infty.
	\end{cases}
\end{equation}
Alternatively, one may heuristically understand these limits by
substituting $\rho^{(l)}=0$ to Eqs.~(\ref{eq:C-rho-correspondence}), (\ref{eq:derivprod-rho-correspondence})
(for $x\rightarrow 0$)
or by plugging the asymptote $\rho^{(l)}\simeq (\kappa l)^{-1}$
into Eq.~(\ref{eq:derivprod-rho-correspondence}) and then resorting to
an arithmetic formula (for $x\rightarrow\infty$)
\begin{equation}
	\lim_{L\rightarrow\infty}\left[\frac{1}{L}\sum_{l=1}^L\prod_{l^{\prime}=l}^{L}\left(1-\frac{2}{l^{\prime}}\right)\right] = \frac{1}{3}.
\end{equation}
Either way, reasonable scaling collapse
(except when $L$ is small, just as expected)
shown in Fig.~\ref{fig:DNNAPTscaling}(b) empirically validates the scaling ansatz (\ref{eq:NTK-ansatz}). 

The ansatz (\ref{eq:NTK-ansatz}), together with the asymptotics~(\ref{eq:NTKscalingfunc}),
suggests that $\kappa L$ is a crucial factor
for the properties of NTK at a critical point. If $\kappa L$ is too small,
the resulting $\Theta_{\mathrm{train}}^{(L)}$ becomes nearly rank-1,
which implies slow training of the network in general.
Conversely, if $\kappa L$ is too large, NTK asymptotically
behaves like an indicator function
\begin{equation}
	\label{eq:NTKasymptotes}
	\Theta^{(L)}(\boldsymbol{x}_1,\boldsymbol{x}_2) \simeq
	\begin{cases}
		q_c^*L & \boldsymbol{x}_1=\boldsymbol{x}_2;\\
		q_c^*L/3 & \text{otherwise},
	\end{cases}
\end{equation}
which seriously deteoriates the network.
For instance, the dynamics of an output $y(\boldsymbol{x})$ for
an input $\boldsymbol{x}$ outside the training dataset under the gradient descent,
namely,
\begin{equation}
	\frac{\mathrm{d}y(\boldsymbol{x};\boldsymbol{\theta}(t))}{\mathrm{d}t} = -\frac{2\eta}{N}\Theta_{\text{test}}^{(L)}(\boldsymbol{x}) \Delta\boldsymbol{y}(\boldsymbol{\theta}(t))
\end{equation}
with
\begin{equation}
	\Theta_{\text{test}}^{(L)}(\boldsymbol{x}) := (\Theta^{(L)}(\boldsymbol{x},\boldsymbol{x}_1),\cdots,\Theta^{(L)}(\boldsymbol{x},\boldsymbol{x}_N)),
\end{equation}
becomes almost independent of $\boldsymbol{x}$,
which indicates a poor generalization performance.
To put it differently, even if initialized at a critical point,
the networks with too small $\kappa L$ behave as if
they were in the ordered phase, whereas
those with too large $\kappa L$ in the chaotic phase
(see also Xiao \textit{et al.}~\cite{Xiao2020}).
Thus, $\kappa L$ should be properly chosen to fully
exploit the benefit of initialization at a critical point.
Along this line of thinking, the \textit{curse of depth} reported
by Hayou \textit{et al.}~\cite{Hayou2022} can be understood
as a devastating consequence of infinite $\kappa L$,
rather than as an intrinsic limitation of the infinitely wide networks.
One caveat is that the range within which
$\kappa L$ should be tuned as suggested from Fig.~\ref{fig:DNNAPTscaling}(b) alone,
namely (below, $\rho^{(0)}_{\mathrm{max}}$ and $\rho^{(0)}_{\mathrm{min}}$ denote the maximum and minimum non-zero cosine distance $\rho^{(0)}$
achieved in a training dataset, respectively)
\begin{equation}
	\label{eq:kappaL-favrange}
	0.1/\omega\rho^{(0)}_{\mathrm{max}} \lesssim \kappa L \lesssim 10/\omega\rho^{(0)}_{\mathrm{min}}
\end{equation}
so that $\Theta^{(L)}(\boldsymbol{x}_i,\boldsymbol{x}_j)$ for
all the pairs of training inputs $\boldsymbol{x}_i,\boldsymbol{x}_j$ ($i\ne j$)
do not fall into the same asymptotic regime, is rather loose.
We might be able to tighten the range by a more thorough analysis of NTK
for a dataset at hand, although we cannot expect such a tighter range
to be carried over different datasets.
We plan to investigate this point more in the near future.

\subsection{Scaling results for the finite networks and different architectures}\label{subsec:scaling-finite-and-arch}
\begin{figure}[tbp]
	\centering
	\includegraphics[width=\columnwidth]{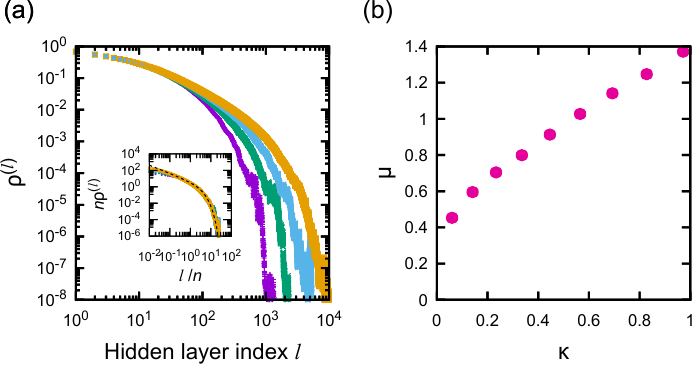}
	\caption{Finite-size scaling of the order-to-chaos transition
		in the multilayer perceptrons~(\ref{eq:FCrecursive}).
		(a) The order parameter $\rho^{(l)}$ (see Eq.~(\ref{eq:NNopdef})) at
		a critical point ($(\sigma_{w;c},\sigma_b)\sim (1.39558,0.3)$ with $\tanh$ activation;
		$\kappa\sim 0.233498$) for
		various widths $n$ (ranging from 50 (purple) to 400 (orange)),
		empirically averaged over $10^4$ independent runs.
		Two orthogonal inputs $\boldsymbol{x}_1,\boldsymbol{x}_2$ 
		(with $\|\boldsymbol{x}_1\|_2=\|\boldsymbol{x}_2\|_2=1$) of
		size $n_{\mathrm{in}}=10$ were given.
		The inset shows the same data rescaled according to
		the univesal scaling ansatz~(\ref{eq:FSS-FC}).
		The black dashed curve indicates the solution~(\ref{eq:FCscalingfunc})
		of the phenomenological description~(\ref{eq:FC-FSphenomenology})
		with $(\kappa,\mu)\sim (0.233498,0.6601)$,
		where $\mu$ was chosen by fitting the solution
		to the empirical result for $n=400$.
		(b) The nonuniversal metric factor $\mu$ as a function of $\kappa$
		in the case of $\tanh$ activation,
		where $\mu$ for each $\kappa$ was estimated from the same fitting as the inset of (a)
		using the empirical $\rho^{(l)}$ for $n=200$.
	}
	\label{fig:DNNAPTfinitewidth}
\end{figure}

Another virtue of the universal scaling laws is that they give us
useful intuition even into the networks of finite width, where
quantitatively tracking the deviation from the Gaussian process
can be cumbersome (if not impossible \cite{Grosvenor2022,Yaida2020}).
To illustrate this point, let us consider the finite-size scaling of the neural network 
at a critical point. Since the order parameter $\rho^{(l)}$
in the mean-field theory is defined through
the Pearson correlation coefficient $c^{(l)}$ (see Eq.~(\ref{eq:MFPearsoncorr})),
definition of the finite-width counterpart is straightforward:
\begin{equation}
	\label{eq:NNopdef}
	\rho^{(l)}:=1-\frac{\sum_i(z_{1;i}^{(l)}-Z_1^{(l)})(z_{2;i}^{(l)}-Z_2^{(l)})}{\sqrt{\sum_i (z_{1;i}^{(l)}-Z_1^{(l)})^2 \sum_i (z_{2;i}^{(l)}-Z_2^{(l)})^2}},
\end{equation}
where $\boldsymbol{z}_1^{(l)},\boldsymbol{z}_2^{(l)}\in \mathbb{R}^n$ are
the preactivations at the $l$-th hidden layer
for different inputs $\boldsymbol{x}_1,\boldsymbol{x}_2$, respectively, $z_{j;i}^{(l)}$ 
the $i$-th element of $\boldsymbol{z}_j^{(l)}$, and $Z_j^{(l)}:=\frac{1}{n}\sum_iz_{j;i}^{(l)}$.
We empirically find that
the order parameter $\rho^{(l)}$ at a critical point
for various widths $n$ admits
the following universal scaling ansatz (Fig.~\ref{fig:DNNAPTfinitewidth}(a)):
\begin{equation}
	\label{eq:FSS-FC}
	\rho^{(l)}[\sigma_w=\sigma_{w;c};n]\simeq n^{-1}f(n^{-1}l).
\end{equation}

The empirical finding above can be heuristically understood
by considering a finite-width correction to the mean-field theory.
In the case of the finite-width networks, the fourth-order (and other even-order)
cumulants come into play,
while the third-order (and other odd-order) ones
vanish just as the odd-order moments do \footnote{This peculiarity explains why 
	the finite size scaling in the multilayer perceptrons is different from that
	in the contact process \cite{Harris1974} on a complete graph,
	where one finds the same $\beta$ and $\nu_{\|}$ (Eq.~(\ref{eq:FCcrit-exponents})) but the exponent
	for finite-size scaling (\ref{eq:FSS-FC}) is replaced
	with $-1/2$. If the third-order cumulant remained non-zero, the leading order for
	the correction would be an order of $n^{-\frac{1}{2}}$.}.
In the spirit of the asymptotic expansion of the probability distribution \cite{Kolassa2006},
this observation indicates that the leading correction to the Gaussian process is
an order of $n^{-1}$, the reciprocal of the width. Thus, together with
an obvious fact that $\rho=0$ is an absorbing state also for the finite networks,
we are led to the following modified phenomenological description:
\begin{equation}
	\label{eq:FC-FSphenomenology}
	\derivwrt{\rho}{l}=-\frac{\mu}{n}\rho-\kappa\rho^2,
\end{equation}
where $\mu$ is a new nonuniversal metric factor.
Unfortunately, theoretical calculation of $\mu$ would be challenging,
since this requires us to analyze the approximate recursion relation up to $O(n^{-1})$
for the covariance, which is no longer closed
within the variance and the covariance
(as opposed to the mean-field theory (\ref{eq:FCMF-q}), (\ref{eq:FCMF-c})).
Still, one can \textit{measure} it by fitting the empirical $\rho^{(l)}$
to the analytical solution of the phenomenological description (\ref{eq:FC-FSphenomenology}):
\begin{equation}
	\label{eq:FCscalingfunc}
	n\rho^{(l)}=\frac{\rho_0\mu}{\rho_0\kappa(e^{\frac{\mu l}{n}}-1)+(\mu/n)e^{\frac{\mu l}{n}}},
\end{equation}
as demonstrated in the inset of Fig.~\ref{fig:DNNAPTfinitewidth}(a).

While it is nowadays well established that
the depth-to-width ratio $L/n$
is a key quantity for describing
the multilayer perceptrons with finite $n$ \cite{Roberts2022,Bahri2024,Hanin2024},
the metric factor $\mu$ enriches this insight by providing
the means to quantitatively characterize the sensitivity of the network to the width.
Specifically, the width $n$ of the network should satisfy
$n\gtrsim \mu L$ so that the signal propagation dynamics therein
is reasonably well approximated by the infinitely wide limit.
This introduces another design consideration for the neural networks.
Recalling the relevance of $\kappa L$ for the training dynamics of the networks,
one would be tempted to use a critical network with larger $\kappa$ to
obtain good generalization with smaller $L$.
However, Fig.~\ref{fig:DNNAPTfinitewidth}(b) empirically suggests that
larger $\kappa$ comes with a cost of larger $\mu$,
which imposes an extra computational burden for larger $n$.
Hence, if one chooses to operate the network near the infinitely wide limit,
one needs to make a trade-off between these two factors for cost-effective training.
Theoretically, a more precise formulation of this idea might be achieved by
studying the finite-width corrections \cite{Huang2020} to the training dynamics,
which is beyond the scope of the present work.

Finally, we briefly discuss the convolutional neural networks
to see how the analogy to absorbing phase transitions
carries over different architectures.
Formally, the recurrence relations for the preactivation
$\boldsymbol{z}^{(l;\alpha)}$
of a $d$-dimensional convolutional neural network (a periodic boundary condition,
also known as circular padding, is assumed for simplicity)
is described as follows (below, $c$ and $k$ denote
the number of channels and the width of the convolution filter, respectively):
\begin{equation}
	\label{eq:Convrecursive}
	\boldsymbol{z}^{(l+1;\alpha)}=\frac{\sigma_w}{\sqrt{ck^d}}\sum_{m=1}^cw^{(l+1;\alpha,m)}\star h(\boldsymbol{z}^{(l;m)})+\sigma_b\boldsymbol{b}^{(l+1;\alpha)},
\end{equation}
where $\star$ denotes the cross-correlation operator
(the summation below is taken over the range $\{(k-1)/2,\cdots,-1,0,1,\cdots,(k-1)/2\}$
for each $j_1,\cdots,j_d$)
\begin{equation}
	(A\star B)_{i_1,\cdots,i_d} := \sum_{j_1,\cdots ,j_d} A_{j_1+\frac{k+1}{2},\cdots,j_d+\frac{k+1}{2}}B_{i_1+j_1,\cdots,i_d+j_d},
\end{equation}
and $h(\boldsymbol{z})$ is a shorthand for element-wise application of $h$ to $\boldsymbol{z}$.
Each element of the convolutional filter $w^{(l;\alpha,m)}\in\mathbb{R}^{k^d}$ and the bias $\boldsymbol{b}^{(l;\alpha)}$
is initialized according to the standard normal distribution $\mathcal{N}(0,1)$.
In the present case, $\rho^{(l)}$ is obtained by first calculating
the correlation coefficient (\ref{eq:NNopdef}) for each channel
and then taking the average over all the channels.

The key difference compared to the multilayer perceptrons is
the locality of the interaction between the neurons.
The neurons within a convolutional layer
interact only locally through the convolutional filters,
in contrast with the multilayer perceptions, where
the network admits the fully connected structure.
The richer dynamics due to the spatial degrees of freedom
can be partly grasped by studying the limit of
infinitely many channels $c\rightarrow\infty$.
In this limit, the neural network is again equivalent to
a Gaussian process~\cite{Garriga-Alonso2018} and the phase diagram of the mean-field theory
remains exactly the same~\cite{Xiao2018}
as the multilayer perceptrons (Fig.~\ref{fig:DNNAPTanalogy}(a)),
making it easy to compare between the two architectures.
Different phases in the convolutional networks are characterized by
how a noise in a single pixel spatially spread in the course of the signal propagation,
in addition to the asymptotic behavior of the order parameter $\rho$.
One can empirically check that the noise eventually decays in the ordered phase, whereas
it spreads ballistically in the chaotic phase.
At a critical point, the spreading process is diffusive,
whose characteristic width $n_*$ scales with the network depth $l$ as $n_*\sim l^{\nu_{\perp}/\nu_{\|}}=\sqrt{l}$, which induces a new critical exponent
\begin{equation}
	\nu_{\perp}=1/2.
\end{equation}
This is exactly what happens in the mean-field theory of
absorbing phase transitions with the spatial degrees of freedom~\cite{Henkel2008}
at a critical point
\begin{equation}
	\label{eq:APTMFwithspatial}
	\frac{\partial\rho}{\partial l}=-\kappa\rho^2+D\nabla^2\rho.
\end{equation}
Thus, the signal propagation dynamics of the convolutional neural networks with
infinitely many channels and input pixels at the critical point is characterized by
two independent metric factors $(\kappa,D)$.
Theoretical calculation of the new metric factor $D$
could perhaps be done using the similar techniques~\cite{Coutinho1997,Fernandez1997}
for studying the dynamics of wavefronts in
coupled map lattices~\cite{Kaneko1984}, although this is substantially
more challenging than $\kappa$. Since an exact and efficient
algorithm to compute NTK is also available for
the convolutional networks~\cite{Arora2019},
it would be interesting to investigate the role of $D$
in the training dynamics. At any rate, we believe it is safe to say that
the analogy to absorbing phase transitions is a promising
insight for studying deep neural networks outside
the multilayer perceptrons.

\begin{figure}[tbp]
	\centering
	\includegraphics[width=\columnwidth]{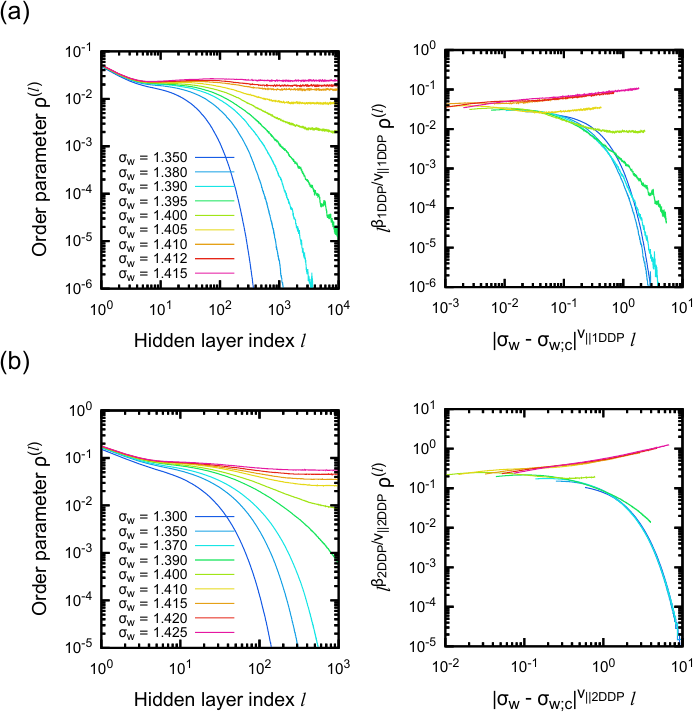}
	\caption{Directed percolation (DP)
		scaling in the order-to-chaos transition 
		in the convolutional neural networks (\ref{eq:Convrecursive}).
		(a)~The order parameter $\rho^{(l)}$ with $d=1$, $n=400$, $k=5$ and $c=10$,
		empirically averaged over $10^6$ independent runs.
		The raw data is shown in the left panel, which is then rescaled according to
		the scaling ansatz (\ref{eq:abs-ansatz}) with the critical exponents
		of $(1+1)$-dimensional DP (\ref{eq:1DDPexponents}) in the right.
		Two orthogonal inputs $\boldsymbol{x}_1,\boldsymbol{x}_2$
		(with $\|\boldsymbol{x}_1\|_2=\|\boldsymbol{x}_2\|_2=1$)
		of size $n$ were given.
		$\sigma_{w;c}:=1.408$ is chosen to find the scaling collapse. 
		(b)~Similar with (a), but with $d=2$, $n=100$, $k=3$ and $c=5$,
		averaged over 4000 independent runs.
		The critical exponents of $(2+1)$-dimensional DP (\ref{eq:2DDPexponents})
		and $\sigma_{w;c}:=1.404$ were used to find the scaling collapse in this case.
		In both (a) and (b), $\tanh$ activation is used and the parameter $\sigma_b$
		for the bias vectors is fixed to be 0.3.
	}
	\label{fig:DNNAPTconv}
\end{figure}

The analogy to absorbing phase transitions also gives us
a nontrivial and yet intuitive insight into
the signal propagation dynamics of the convolutional neural networks with finite channels $c$,
although the implications to the training dynamics may be
less direct, just as in the case of the finite-width multilayer perceptrons.
Empirical evidence we show in Fig.~\ref{fig:DNNAPTconv} suggests
the following phenomenology.
If $c$ is finite, the dynamics of the covariance
(and hence of the order parameter $\rho^{(l)}$) is
no longer deterministic but is accompanied by
a multiplicative noise, whose amplitude is
asymptotically proportional to $\sqrt{\rho^{(l)}}$
as normally expected for models
with microscopic stochastic elements \cite{Grinstein1997}.
The noise works as a relevant perturbation to
the mean-field theory~(\ref{eq:APTMFwithspatial})
in the sense of the renormalization group,
and it changes the asymptotic scaling behavior of the network at a large scale
to that of the directed percolation (DP) universality class~\cite{Broadbent1957}.
As such, the universal scaling ansatz (\ref{eq:abs-ansatz})
remains valid with different critical exponents.
For instance, reasonable scaling collapse can be found
for the order parameter $\rho^{(l)}$ in the spatially one-dimensional
convolutional networks (Fig.~\ref{fig:DNNAPTconv}(a)) using
the exponents for the $(1+1)$-dimensional DP universality class \cite{Jensen1999}
\begin{equation}
	\label{eq:1DDPexponents}
	\beta_{\mathrm{1DDP}}\sim 0.27649,\quad \nu_{\|\mathrm{1DDP}}\sim 1.73385,
\end{equation}
although the order parameter can saturate
to a non-zero value slightly below the critical point
(causing the tilts in the rescaled plot)
mainly because the perfect order ($\rho^{(l)} = 0$) is
virtually unachievable;
the imperfection acts as a small fluctuation,
to which the setups near the critical point is particularly sensitive
\footnote{Other possible (albeit relatively minor) reasons for the deviation from the scaling collapse
	include the finite-size effect and deviation of the noise from
	the asymptotic behavior
	(which is particularly true if the number $c$ of channels is small).}.
We also checked that the essentially same scenario holds
true for the two-dimensional convolutional networks (Fig.~\ref{fig:DNNAPTconv}(b)),
where the critical exponents $\beta,\nu_{\|}$ are replaced~\cite{Voigt1997,Wang2013} with
\begin{equation}
	\label{eq:2DDPexponents}
	\beta_{\mathrm{2DDP}}\sim 0.58,\quad \nu_{\|\mathrm{2DDP}}\sim 1.29.
\end{equation}
Remarkably, the deviation from the scaling collapse is less prominent
in the two-dimensional networks than in their one-dimensional counterparts,
which is consistent with our interpretation of the tilts since
fluctuations generally become less relevant
as the spatial dimensionality goes up~\cite{Henkel2008}.

The correspondence to the DP universality class suggest that
the signal propagation dynamics in the convolutional networks
is highly nontrivial, especially given the notorious difficulty of
exactly solving DP~\cite{Guttmann2000}.
Yet, thanks to the universality of the scaling laws of absorbing phase transitions,
semi-quantitative predictions can be gained via simple phenomenological considerations.
In particular, the most informative combination of the width $n$ and the depth $L$
for describing the behavior of the critically initialized deep convolutional networks
may be changed from $L/n$ to $L/n^{\nu_{\|}/\nu_{\perp}}$ using the corresponding
critical exponents
\begin{equation}
	\nu_{\perp\text{1DDP}}\sim 1.096854,\quad \nu_{\perp\text{2DDP}}\sim 0.73.
\end{equation}

\section{Discussion}\label{sect:discussion}
To summarize, we pursued the analogy between the behavior
of the conventional deep neural networks
and absorbing phase transitions in the present work.
During the pursuit, we demonstrated that the signal propagation dynamics
in the untrained neural networks follows the universal scaling laws, while
the specific details are summarized using the associated nonuniversal metric factors.
In particular, the nonuniversal metric factor $\kappa$ was shown to
play a significant role in the training dynamics of the multilayer perceptrons:
its product with the network depth $L$ should be tuned for optimal generalization.
Thus, the present work provides useful insights
into the neural networks with many but finite hidden layers,
which complements our understanding of
two-layer \cite{Chen2020,Ju2021} or
infinitely-deep networks \cite{Xiao2020,Hayou2022}.
The framework can be readily extended to
ReLU-like activation functions (albeit with different exponents),
which consequently underlines the significance of
properly choosing the amount of leak;
see Appendix~\ref{sect:scale-invariant-activation}.
Furthermore, we provided numerical evidence suggesting that
the analogy to absorbing phase transitions well captures
the signal propagation dynamics in the neural networks
with finite width or different architecture,
holding great promise for future developments.

Let us emphasize that successful deep learning can only be
achieved via a complicated interplay among various setups,
even in one of the simplest cases where NTK describes
the training dynamics reasonably well.
In addition to the relevance of
initialization at criticality \cite{Schoenholz2017, Hayou2022} and
of proper scaling of a learning rate with respect to depth \cite{Xiao2020},
the present work demonstrates the necessity of
a more specific, depth-dependent choice of hyperparameters.
Furthermore, in the case of a large but finite width,
one should also strike a balance between width and depth for efficiency.
The fact that all these setups need to be considered simultaneously
highlights the major challenge in deep learning,
which necessitates extensive study
on hyperparameter optimization \cite{Yu2020}.
To put it the other way around,
considerable theoretical insight into
the mechanism behind the recent success of deep learning may be obtained
by studying the neural networks near the realm of NTK,
contrary to a common belief \cite{Chizat2019, Nichani2022}.

In a broader context, the present work hopefully exemplifies
a subtle relationship between criticality and intelligence.
In the case of the artificial neural networks,
being at criticality alone is not sufficient for successful learning,
although it is likely to be necessary.
Intriguingly, this theoretical insight is consistent with the experimental findings
on real neural systems: while deviation from criticality
often results in an altered or abnormal state of consciousness \cite{Gervais2023,Xu2024},
a sign of criticality does not necessarily imply presumable capability of 
performing intellectual tasks \cite{Friedman2012,Shew2015,Kosmidis2018,Fontenele2019}.

Then, what can we learn from the present work to
improve our understanding of intelligence in living systems?
One lesson may be that we should pay closer attention to
nonuniversal aspects of the critical dynamics,
particularly in light of \textit{memory effects}.
A system tuned at criticality exhibits a macroscopic memory effect
due to the divergent correlation time \cite{Janssen1989,Zheng2000}.
This is the physical origin of the dynamic scaling (\ref{eq:initial-slip-ansatz})
of the critical initial slip.
The rate of memory loss is nonuniversal:
in the multilayer perceptrons, for instance,
the metric factor $\kappa$ characterizes the memory loss per hidden layer,
and the value (\ref{eq:kappa-formula}) is \textit{not} shared
among the different points on the phase boundary,
unlike the critical exponents (\ref{eq:FCcrit-exponents}).
The presence of the favorable range (\ref{eq:kappaL-favrange}) for $\kappa L$
suggests that the memory characteristics of the neural networks
may play a key role in their intellectual property,
whose quantification goes beyond the measurement of the exponents.
We hope that the present work inspires the development of techniques to
quantitatively characterize nonuniversal features of
critical states in real neural systems.

We foresee some interesting directions for future work.
Apart from the ones mentioned in the previous Section
\footnote{Namely, (a) derivation of more precise design principles
	of the neural networks from the universal scaling laws,
	possibly with prior knowledge about
	a dataset at hand (Sect.~\ref{subsec:infwide-scaling})
	and (b) uncovering the details of the trade-off
	between the width and the depth (Sect.~\ref{subsec:scaling-finite-and-arch}).},
one of the most natural directions is to extend the present framework
to more modern architectures \cite{Bertschinger2004,Takasu2024,Yang2017}.
In particular, the skip connections
employed in the residual neural networks (ResNet \cite{He2016})
may be seen as a stimulus kicking the system out of an absorbing state:
even if the signals are collapsed during the propagation within a residual block,
the skip connection breaks the order before entering the new block.
Given that a combination of external drive and dissipation into absorbing states
has been conjectured to be a key ingredient
of self-organized criticality in physical systems
\cite{Munoz2018,Girardi-Schappo2021}, the skip connections may have
their unique benefits in deep learning as well.
Another, albeit less straightforward, direction is to
provide thermodynamic foundations of
the speed-accuracy trade-off in deep learning.
The deep neural networks in the ordered phase
sacrifice speed for accuracy, and vice versa in the chaotic phase.
Since the speed-accuracy trade-off has been
extensively studied in the context of
living systems \cite{Lan2012,Rao2015,Banerjee2017,Ouldridge2017},
thermodynamical insights developed therein
are likely to be helpful (and indeed, very recently, the trade-off in the diffusion models
has been studied from a thermodynamic viewpoint \cite{Ikeda2024}),
although pursuing this direction would call for an improved understanding of
the thermodynamics of absorbing phase transitions \cite{Zeraati2012,Harada2019}.
We believe that further investigation into a parallel between
intelligence in living systems and that in artificial neural networks
will lead us to a lot of exciting developments, beneficial for
both physics and machine learning communities.

\appendix
\section{Derivation of the critical exponents for the multilayer perceptrons}
\label{sect:FCexponentderivation}
Here we derive the critical exponents (\ref{eq:FCcrit-exponents}) 
of the multilayer perceptrons in the main text. 
That is, we show that the order parameter $\rho^*$ (see Eq.~(\ref{eq:rho-c-correspondence}))
at the fixed point of the iterative $\mathcal{C}$-map (\ref{eq:iterCmap})
and the correlation depth $\xi_{\|}$ as defined by Eq.~(\ref{eq:FCcorrdepthdef})
respectively exhibits linear onset in the vicinity of the critical point
(although we have already seen it empirically in Fig.~\ref{fig:DNNAPTanalogy}(c)).
To achieve this goal, we expand the mean-field theory (\ref{eq:FCMF-q}), (\ref{eq:FCMF-c}) with respect to
infinitesimally small deviation $\delta\sigma_w$ from the critical point $\sigma_{w;c}$.

First, let us show the continuity of $q^*$ as a function of $\sigma_w$
at the edge of chaos for later convenience.
Consider the fixed point $q^*$ of the mean-field theory (\ref{eq:FCMF-q}) for infinitesimally different $\sigma_w$,
and let $\delta\sigma_w$ and $\delta q^*$ respectively denote the increment in $\sigma_w$ and $q^*$.
Then, we compare the equality for the fixed point of $q$ as follows:
\begin{equation}
	\label{eq:qexpand_ref}
	q^* = \sigma_w^2\int\mathcal{D}z\,h^2(\sqrt{q^*}z)+\sigma_b^2;
\end{equation}
\begin{equation}
	\label{eq:qexpand_infinitesimal}
	\begin{array}{rcl}
		q^*+\delta q^* & = & \displaystyle (\sigma_w+\delta\sigma_w)^2\int\mathcal{D}z\,h^2(\sqrt{q^*+\delta q^*}z)+\sigma_b^2\\
		& \simeq & \displaystyle \sigma_w^2\int\mathcal{D}z\,h^2(\sqrt{q^*}z)+\sigma_b^2\\
		& & \displaystyle \mbox{} + \delta q^*\,\sigma_w^2\int\mathcal{D}z\,\frac{z}{\sqrt{q^*}}h(\sqrt{q^*}z)h^{\prime}(\sqrt{q^*}z)\\
		& & \displaystyle \mbox{} + 2\delta\sigma_w \, \sigma_w\int\mathcal{D}z\,h^2(\sqrt{\qfixedpt{}}z),
	\end{array}
\end{equation}
where we have neglected difference of $O((\delta q^*)^2)$, $O((\delta\sigma_w)^2)$ or $O(\delta q^*\delta\sigma_w)$.
By subtracting Eq.~(\ref{eq:qexpand_ref}) from Eq.~(\ref{eq:qexpand_infinitesimal}),
we find
\begin{equation}
	\label{eq:qstarderiv}
	\begin{array}{rcl}
		\delta q^* & \simeq & \displaystyle \frac{\displaystyle 2\sigma_w\int\mathcal{D}z\,h^2(\sqrt{\qfixedpt{}}z)}{\displaystyle 1-\sigma_w^2\int\mathcal{D}z\,\frac{z}{\sqrt{q^*}}h(\sqrt{q^*}z)h^{\prime}(\sqrt{\qfixedpt{}}z)}\delta\sigma_w\\
		& =: & \alpha\delta\sigma_w.
	\end{array}
\end{equation}
In particular at the critical point $\sigma_{w;c}$, the coefficient $\alpha$ can be further simplified to
\begin{equation}
	\alpha=\frac{\displaystyle 2\int\mathcal{D}z\,h^2(\sqrt{q^*_c}z)}{\displaystyle -\sigma_{w;c}\int\mathcal{D}z\,h(\sqrt{q^*_c}z)h^{\prime\prime}(\sqrt{q^*_c}z)},
\end{equation}
where $q^*_c$ is the fixed point of Eq.~(\ref{eq:FCMF-q}) at the edge of chaos.
It turns out that the numerator and the denominator of the RHS of Eq.~(\ref{eq:qstarderiv}) converge
to a finite value, so does $\alpha$ itself.

Next, we study the behavior of $\xi_{\|}^{-1}$ slightly below the critical point. To do this, 
we expand Eq.~(\ref{eq:infFCcorrdepth}) with respect to 
an infinitesimal deviation $\delta \sigma_w$ from the critical point $\sigma_{w;c}$:
\begin{equation}
	\label{eq:xionset_subcrit}
	\begin{array}{l}
		e^{-\frac{1}{\xi_{\|}(\sigma_{w;c}-\delta\sigma_w)}}\\ 
		= \displaystyle (\sigma_{w;c}-\delta\sigma_w)^2\int\mathcal{D}z\,h^{\prime 2}(\sqrt{q_c^*-\alpha\delta\sigma_w}z)\\
		\displaystyle \simeq 1-\left[\frac{2}{\sigma_{w;c}}+\frac{\alpha\sigma_{w;c}^2}{\sqrt{q_c^*}}\int\mathcal{D}z\,zh^{\prime}(\sqrt{q_c^*}z)h^{\prime\prime}(\sqrt{q_c^*}z)\right]\delta\sigma_w\\
		= \displaystyle 1-\gamma_1\delta\sigma_w,\\
	\end{array}
\end{equation}
where
\begin{equation}
	\gamma_1:=\frac{2}{\sigma_{w;c}}\left(1-\frac{\displaystyle (q_c^*-\sigma_b^2)\int\mathcal{D}z\,zh^{\prime}(\sqrt{q_c^*}z)h^{\prime\prime}(\sqrt{q_c^*}z)}{\displaystyle \sqrt{q_c^*}\int\mathcal{D}z\,h(\sqrt{q^*_c}z)h^{\prime\prime}(\sqrt{q^*_c}z)}\right).
\end{equation}
The coefficient $\gamma_1$ remains finite for
activation functions in the $K^*=0$ universality class,
and hence $\xi_{\|}^{-1}$ decreases to 0 as $\sigma_{w}\uparrow \sigma_{w;c}$ in an asymptotically
linear manner.

The correlation depth $\xi_{\|}^{-1}$ slightly above the critical point (see Eq.~(\ref{eq:auxUdef}) for the definitions of $u_1^*, u_2^*$)
\begin{equation}
	\begin{array}{l}
		\label{eq:corrdep-abovecrit}
		e^{-\frac{1}{\xi_{\|}(\sigma_{w;c}+\delta\sigma_w)}}\\
		= \displaystyle (\sigma_{w;c}+\delta\sigma_w)^2\int\mathcal{D}z_1\int\mathcal{D}z_2\,h^{\prime}(u_1^*+\delta u_1^*)h^{\prime}(u_2^*+\delta u_2^*)\\
	\end{array}
\end{equation}
can be studied similarly, but we need first to analyze
the behavior of the fixed point $c^*$ of the iterative $\mathcal{C}$-map (\ref{eq:iterCmap})
as a function of $\sigma_w$, due to the $c$-dependence of $u_2$.
Hence we expand the $\mathcal{C}$-map slightly above the critical point
(that is, $\sigma_w=\sigma_{w;c}+\delta\sigma_w$)
around the trivial fixed point $c^{(l)}=1$
\begin{equation}
	\label{eq:cmap-expansion}
	\begin{array}{rcl}
		c^{(l+1)}-c^{(l)} & = & \displaystyle \left(\left.\frac{\mathrm{d} c^{(l+1)}}{\mathrm{d} c^{(l)}}\right|_{c^{(l)=1}}-1\right)(c^{(l)}-1)\\
		& & \displaystyle \mbox{}+\frac{1}{2}\left.\frac{\mathrm{d}^2 c^{(l+1)}}{\mathrm{d} c^{(l)2}}\right|_{c^{(l)}=1}(c^{(l)}-1)^2+\cdots.
	\end{array}
\end{equation} 
Notice that essentially the same calculation as the one
for analyzing linear stability of the trivial fixed point \cite{Schoenholz2017}
can be repeated to inductively see
\begin{equation}
	\left.\frac{\mathrm{d}^nc^{(l+1)}}{\mathrm{d}c^{(l)n}}\right|_{c^{(l)}=1}=\sigma_w^2q^{*n-1}\int\mathcal{D}z \left(\frac{\mathrm{d}^nh}{\mathrm{d}z^n}(\sqrt{q^*}z)\right)^2,
\end{equation}
which implies these derivatives are positive and finite at any order. Particularly in the vicinity of the critical point, we have, from Eq.~(\ref{eq:xionset_subcrit}),
\begin{equation}
	\left.\frac{\mathrm{d}c^{(l+1)}}{\mathrm{d}c^{(l)}}\right|_{c^{(l)}=1}-1=\gamma_1\delta\sigma_w+o(\delta\sigma_w).
\end{equation}
By taking the first two terms of the expansion (\ref{eq:cmap-expansion}) into account
and solving it with respect to $\delta\rho:=1-c^*$ at the fixed point,
one can see that the leading contribution for $\delta\rho$ is of order $\delta\sigma_w$,
more specifically
\begin{equation}
	\label{eq:zeta-expression}
	\delta\rho=\gamma_1 \frac{\displaystyle 2\int\mathcal{D}z\,h^{\prime 2}(\sqrt{q^*_c}z)}{\displaystyle q^*_c\int\mathcal{D}z\,h^{\prime\prime 2}(\sqrt{q^*_c}z)}\delta\sigma_w =: \zeta\,\delta\sigma_w.
\end{equation}
This result implies that the critical exponent $\beta$ associated with the onset of the order parameter is $1$.

Now we are in the position of studying $\xi_{\|}^{-1}$ slightly above the critical point (\ref{eq:corrdep-abovecrit}):
\begin{equation}
	\label{eq:xionset_supercrit}
	\begin{split}
		&1-e^{-\frac{1}{\xi_{\|}(\sigma_{w;c}+\delta\sigma_w)}}\\
		&\displaystyle \simeq \left[\zeta\sigma_{w;c}^2\sqrt{q_c^*}\left(\int\mathcal{D}z\,zh^{\prime}(\sqrt{q_c^*}z)h^{\prime\prime}(\sqrt{q_c^*}z) \right.\right.\\
		&\qquad\displaystyle \left.-\int\mathcal{D}z_1\int\mathcal{D}z_2 \sqrt{q_c^*}z_2^2h^{\prime}(\sqrt{q_c^*}z_1)h^{\prime\prime\prime}(\sqrt{q_c^*}z_1)\right)\\
		&\qquad\displaystyle \left.-\frac{\alpha\sigma_{w;c}^2}{\sqrt{q_c^*}}\int\mathcal{D}z\,zh^{\prime}(\sqrt{q_c^*}z)h^{\prime\prime}(\sqrt{q_c^*}z)-\frac{2}{\sigma_{w;c}}\right]\delta\sigma_w\\
		&= \displaystyle \gamma_2\,\delta\sigma_w,
	\end{split}
\end{equation}
where
\begin{equation}
	\begin{array}{rcl}
		\gamma_2 & := & \displaystyle \zeta \frac{\displaystyle q_c^* \int\mathcal{D}z\,h^{\prime\prime 2}(\sqrt{q_c^*}z)}{\displaystyle \int\mathcal{D}z\,h^{\prime 2}(\sqrt{q_c^*}z)}-\gamma_1\\
		& = & \gamma_1=:\gamma.
	\end{array}
\end{equation}
This indicates that $\xi_{\|}^{-1}$ decreases to 0 as $\sigma_{w}\downarrow\sigma_{w;c}$
in an asymptotically linear manner. Thus, it is confirmed that $\nu_{\|}=1$. Note that the contribution of order $\delta\sigma_w^{\frac{1}{2}}$ vanishes because
\begin{equation}
	\int_{-\infty}^{\infty}\mathrm{d}z\frac{z}{\sqrt{2\pi}}e^{-\frac{z^2}{2}}= 0.
\end{equation}

Although the main purpose of this Appendix, namely the derivation of the critical exponents $\beta,\nu_{\|}$,
has already been completed, let us discuss the nonuniversal metric factors $\gamma,\kappa$
introduced in Eq.~(\ref{eq:abs-ansatz}).
By comparing the results (\ref{eq:xionset_subcrit}), (\ref{eq:zeta-expression}), (\ref{eq:xionset_supercrit})
with the solution of the mean-field theory of absorbing phase transition \cite{Henkel2008}
\begin{equation}
	\frac{\mathrm{d}\rho}{\mathrm{d}t} = \gamma_{\leftrightarrow}(\sigma_w-\sigma_{w;c})\rho-\kappa\rho^2,
\end{equation}
we find the following results:
\begin{equation}
	\begin{array}{l}
		\gamma_{\leftrightarrow} = \gamma\\
		\displaystyle = \frac{2}{\sigma_{w;c}}\left(1-\frac{\displaystyle (q_c^*-\sigma_b^2)\int\mathcal{D}z\,zh^{\prime}(\sqrt{q_c^*}z)h^{\prime\prime}(\sqrt{q_c^*}z)}{\displaystyle \sqrt{q_c^*}\int\mathcal{D}z\,h(\sqrt{q^*_c}z)h^{\prime\prime}(\sqrt{q^*_c}z)}\right);
	\end{array}
\end{equation}
\begin{equation}
	\begin{array}{rcl}
		\kappa & = & \gamma/\zeta\\
		& = & \displaystyle \frac{\displaystyle q^*_c\int\mathcal{D}z\,h^{\prime\prime 2}(\sqrt{q^*_c}z)}{\displaystyle 2\int\mathcal{D}z\,h^{\prime 2}(\sqrt{q^*_c}z)}.\\
	\end{array}
\end{equation}
By repeating the same argument for a fixed $\sigma_w > (h^{\prime}(0))^{-1}$,
one can arrive at the same critical exponents with 
the different metric factor $\gamma_{\updownarrow}$;
see Eq.~(\ref{eq:metricgamma-fixedw}).

\section{Scale-invariant activation functions}\label{sect:scale-invariant-activation}

\begin{figure*}[t]
	\centering
	\includegraphics[width=\textwidth]{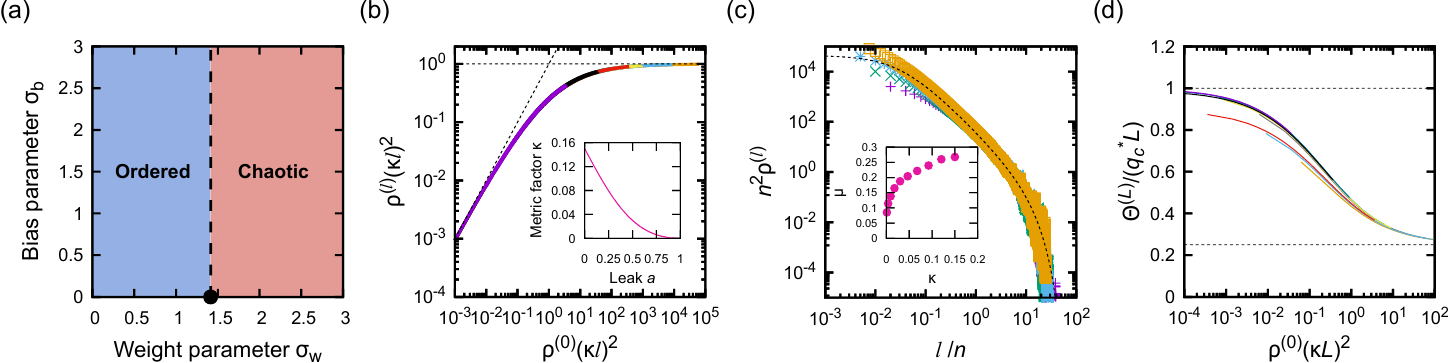}
	\caption{Universal scaling in the infinitely-wide multilayer perceptrons (\ref{eq:FCrecursive})
		with scale-invariant activation functions (\ref{eq:actf-definition-SI}).
		(a)~The phase diagram of the signal propagation for ReLU activation ($a=0$).
		A dashed line is used to indicate the phase boundary because the preactivation variance
		diverges at the boundary unless $\sigma_b = 0$ (a black point).
		(b)~The order parameter $\rho^{(l)}$ at the edge of chaos (\ref{eq:scale-invariant-eoc}) 
		for various combinations of cosine distances $\rho^{(0)}$
		and leak parameters $a$
		(the resulting $\rho^{(0)}\kappa^2$ ranges from $3.6\times 10^{-8}$ (purple) to $6.3\times 10^{-4}$ (orange)), 
		calculated from the mean-field theory (\ref{eq:FCMF-q}), (\ref{eq:FCMF-c})
		and then rescaled according to the scaling ansatz (\ref{eq:scale-invariant-ansatz}).
		The dashed lines are guides-to-eye for the asymptotic behavior (\ref{eq:slipg-asymptotes})
		of the scaling function $g$.
		The inset shows the nonuniversal
		metric factor $\kappa$ as a function of $a$.
		(c)~The main panel shows the order parameter $\rho^{(l)}$ (see Eq.~(\ref{eq:NNopdef})) at
		a critical point ($(\sigma_{w;c},\sigma_b)\sim (\sqrt{2},0)$ with ReLU activation;
		$\kappa = \sqrt{2}/(3\pi)\sim 0.150053$) for
		various widths $n$ (ranging from 50 (purple) to 400 (orange)),
		empirically averaged over $4\times 10^4$ independent runs
		and rescaled according to the universal scaling ansatz (\ref{eq:FSS-ReLUFC}).
		Two orthogonal inputs $\boldsymbol{x}_1,\boldsymbol{x}_2$ 
		(with $\|\boldsymbol{x}_1\|_2=\|\boldsymbol{x}_2\|_2=1$) of
		size $n_{\mathrm{in}}=10$ were given.
		The black dashed curve indicates the solution~(\ref{eq:ReLUFCscalingfunc})
		of the phenomenological description~(\ref{eq:ReLUFC-FSphenomenology})
		with $(\kappa,\mu)\sim (0.150053,0.2676)$,
		where $\mu$ was chosen by fitting the solution
		to the empirical result for $n=200$.
		The inset shows the nonuniversal metric factor $\mu$ as a function of $\kappa$
		in the case of scale-invariant activation functions,
		where $\mu$ for each $\kappa$ was estimated from the same fitting as the main panel
		using the empirical $\rho^{(l)}$ for $n=200$.
		(d) The NTK $\Theta^{(L)}(\boldsymbol{x}_1,\boldsymbol{x}_2)$ (see Eq.~(\ref{eq:FC-NTK})) for various network depths,
		rescaled according to the universal scaling ansatz (\ref{eq:ReLUs-NTK-ansatz}).
		The same combinations of $(\rho^{(0)},a)$ as (b) were used.
		The two horizontal lines are guides-to-eye for
		the asymptotic behavior as $L\rightarrow\infty$ \cite{Hayou2022}:
		the upper one for $\boldsymbol{x}_1=\boldsymbol{x}_2$ ($\rho^{(0)}=0$),
		the lower one for $\boldsymbol{x}_1\ne\boldsymbol{x}_2$ ($\rho^{(0)}>0$).
	}
	\label{fig:scale-invariant-activation}
\end{figure*}

The purpose of this Appendix is to study the signal propagation dynamics
of the infinitely wide multilayer perceptrons with scale-invariant activation functions
(in the following, $a$ is a non-negative parameter often referred to as \textit{leak})
\begin{equation}
	\label{eq:actf-definition-SI}
	h(x)=
	\begin{cases}
		ax & x<0;\\
		x & x\ge 0.
	\end{cases}
\end{equation}
The order-to-chaos transition in the neural networks of this kind
is slightly different from the one discussed in the main text:
qualitative change of the behavior can be found
in the variance $q^{(l)}$ rather than in the covariance $C^{(l)}$.
One can see, by carrying out the integration in Eq.~(\ref{eq:FCMF-q}),
that $q^{(l)}$ exhibits the transition between convergence to some constant $q^*$
and exponential divergence at $\sigma_w=\sigma_{w;c}:=\sqrt{2/(1+a^2)}$,
regardless of $\sigma_b$ \cite{Hayou2019}.
In particular, $q^{(l)}$ stays constant throughout the network if
\begin{equation}
	\label{eq:scale-invariant-eoc}
	(\sigma_w,\sigma_b)=\left(\sqrt{\frac{2}{1+a^2}},0\right),
\end{equation}
while it diverges linearly if $\sigma_b\ne 0$ (Fig.~\ref{fig:scale-invariant-activation}(a)).
Note that a special case of Eq.~(\ref{eq:scale-invariant-eoc}) for $a=0$ is nothing but
the well-known He initialization \cite{He2015}
for ReLU activation. With this initialization scheme,
we have a well-defined iterative $\mathcal{C}$-map (\ref{eq:iterCmap})
and hence we can study the order parameter $\rho^{(l)}$ defined in Eq.~(\ref{eq:rho-c-correspondence});
Cho and Saul \cite{Cho2009} provide technical details on how to analytically
deal with the integration appearing in Eq.~(\ref{eq:iterCmap}) in the case of ReLU activation.
Notice also that $q^*$ vanishes for $\sigma_w<\sigma_{w;c}$ if $\sigma_b=0$.
In other words, the neurons within sufficiently deep
hidden layers \textit{die out} \cite{Lu2020},  
which is reminiscent of an absorbing phase transition discussed in the main text.

A natural question is whether one can apply
the universal scaling of absorbing phase transitions
in the present case, which we will address below.
As visualized in Fig.~\ref{fig:scale-invariant-activation}(b),
we find that the order parameter $\rho^{(l)}$ at the edge of chaos (\ref{eq:scale-invariant-eoc})
follows the universal scaling ansatz
\begin{equation}
	\label{eq:scale-invariant-ansatz}
	\rho^{(l)}\simeq (\kappa l)^{-2}g(\rho^{(0)}(\kappa l)^2),\quad \kappa = \frac{\sqrt{2}(1-a)^2}{3(1+a^2)\pi}
\end{equation}
with a suitable scaling function $g$ having the same asymptotic behavior as Eq.~(\ref{eq:slipg-asymptotes}),
where we dropped the metric factor $\omega$ associated with an initial condition
because $\omega=1$ in this case.
The difference in the scaling exponent compared to the result (\ref{eq:initial-slip-ansatz})
in the main text stems from the second-dominant term 
in the iterative $\mathcal{C}$-map (\ref{eq:iterCmap}). Specifically, one can see that
the difference of $\rho^{(l)}$ in the adjacent layers is asymptotically of $\rho^{(l)\frac{3}{2}}$,
rather than $\rho^{(l)2}$:
\begin{equation}
	\begin{array}{rcl}
		\rho^{(l+1)}-\rho^{(l)} & = & \displaystyle -\frac{\sigma_{w;c}^2(1-a)^2}{2\pi}\left(\sqrt{1-c^{(l)2}}-c^{(l)}\cos^{-1}c^{(l)}\right)\\
		& = & \displaystyle -\frac{2\sqrt{2}(1-a)^2}{3(1+a^2)\pi}\rho^{(l)\frac{3}{2}} + O(\rho^{(l)\frac{5}{2}}).\\
	\end{array}
\end{equation}
Since the argument for a finite-width correction
(see just above Eq.~(\ref{eq:FC-FSphenomenology}))
is not sensitive to a selection of the activation function,
the correction can be made in the same manner as
the $K^*=0$ activation functions
\begin{equation}
	\label{eq:ReLUFC-FSphenomenology}
	\frac{\mathrm{d}\rho}{\mathrm{d}l} = -\frac{\mu}{n}\rho-2\kappa\rho^{\frac{3}{2}},
\end{equation}
although the resulting $l$-dependence of the order parameter $\rho^{(l)}$ and
the exponents for the finite-size scaling are different:
\begin{equation}
	\label{eq:ReLUFCscalingfunc}
	n^2\rho^{(l)} \simeq \frac{\mu^2\rho_0}{\left(2\kappa\sqrt{\rho_0}(e^{\frac{\mu l}{2n}}-1)+(\mu/n)e^{\frac{\mu l}{2n}}\right)^2},
\end{equation}
\begin{equation}
	\label{eq:FSS-ReLUFC}
	\rho^{(l)}\simeq n^{-2}f(n^{-1}l).
\end{equation}
Numerical results shown in Fig.~\ref{fig:scale-invariant-activation}(c) validate
the finite-size scaling ansatz (\ref{eq:FSS-ReLUFC}) and suggest that
the positive correlation between the two metric factors $\kappa,\mu$ holds true
for scale-invariant activation functions (see the inset).
Similarly, the universal scaling for NTK holds
for small $\rho^{(0)}$ or large $L$ (Fig.~\ref{fig:scale-invariant-activation}(d)):
\begin{equation}
	\label{eq:ReLUs-NTK-ansatz}
	\Theta^{(L)}(\boldsymbol{x}_1,\boldsymbol{x}_2) \simeq q_c^{*}L\tilde{g}(\rho^{(0)}(\kappa L)^2).
\end{equation}
Thus, essentially the same scenario as the $K^*=0$ activation functions
is applicable to the scale-invariant functions, with a suitable change of the scaling exponents.

The analysis above potentially provides theoretical foundations of
some empirical insights in the literature.
First, we can correctly anticipate that a small leak of $a=0.01$,
commonly referred to as leaky ReLU (LReLU) in the literature, is unlikely to 
have a significant impact on the network performance \cite{Maas2013,Dubey2022},
since the metric factor $\kappa$ changes only by 2\%.
With larger $a$, however, $\kappa$ noticeably decreases
(for instance, it becomes half of the original ReLU at $a=2-\sqrt{3}\sim 0.27$)
and the optimal depth for training increases in a reciprocal manner
(see the final paragraph of Sect.~\ref{subsec:infwide-scaling}).
Consequently, it is possible that the networks with suitably chosen leak 
works better for a fixed task and other network structures,
in particular when the original ReLU network tends to overfit the training data;
the superior performance of very leaky ($a\sim 0.18$) ReLU
reported by Xu \textit{et al.}~\cite{Xu2015} may be seen
as a remarkable manifestation of such phenomenology, 
although the differences in the network architecture must be taken into account for
a direct comparison.




\begin{acknowledgments}
The numerical experiments for supporting our arguments in this work
were performed
on the cluster machine provided by the Institute for Physics of Intelligence, The University of Tokyo.
This work was supported by the Center of Innovations for Sustainable Quantum AI (JST Grant Number JPMJPF2221) and JSPS KAKENHI Grant Number JP23H03818.
T.O.\ and S.T.\ wish to thank support by
the Endowed Project for Quantum Software Research and Education, The University of Tokyo (https://qsw.phys.s.u-tokyo.ac.jp/).
\end{acknowledgments}

%

\end{document}